\pdfoutput=1
\documentclass[11pt]{article}
\usepackage[final]{acl}
\usepackage{amsmath}
\usepackage{times}
\usepackage{latexsym}
\usepackage{multirow} 
\usepackage{xcolor}
\usepackage{hyperref}
\usepackage{url}
\usepackage{graphicx}
\usepackage{multirow}
\usepackage{caption}
\usepackage{subcaption}
\usepackage{tabularx}  
\usepackage{caption}
\usepackage[T1]{fontenc}
\usepackage{booktabs} 
\usepackage[utf8]{inputenc}
\usepackage{microtype}
\usepackage{inconsolata}
\usepackage{graphicx}

\title{DeKeyNLU: Enhancing Natural Language to SQL Generation through Task Decomposition and Keyword Extraction}
 
\author {Jian Chen\textsuperscript{\rm 1 2},
    Zhenyan Chen\textsuperscript{\rm 3},
    Xuming Hu\textsuperscript{\rm 1},
    Peilin Zhou\textsuperscript{\rm 1},
    Yining Hua\textsuperscript{\rm 4},
    Han Fang\textsuperscript{\rm 2}, \\
    \textbf{Cissy Hing Yee Choy}\textsuperscript{\rm 5}, 
    \bf Xinmei Ke\textsuperscript{\rm 2},
    \bf Jingfeng Luo\textsuperscript{\rm 2},
    \bf Zixuan Yuan\textsuperscript{\rm 1}\thanks{Corresponding Author}
\\
    \textsuperscript{\rm 1}The Hong Kong University of Science and Technology (Guangzhou)
    \textsuperscript{\rm 2}HSBC \\
    \textsuperscript{\rm 3}South China University of Technology
    \textsuperscript{\rm 4}Harvard University
    \textsuperscript{\rm 5}Chicago University\\
\text{\{jchen524, pzhou460\}@connect.hkust-gz.edu.cn}, \\
\text{202164700035@mail.scut.edu.cn, yininghua@g.harvard.edu},\\
\text{\{alex.j.chen, jerry.h.fang, shiny.x.m.ke, roger.j.f.luo\}@hsbc.com}, \\ 
\text{hychoy@uchicago.edu, {\{xuminghu, zixuanyuan\}}@hkust-gz.edu.cn}
}

\begin{document}
\maketitle
\begin{abstract}
Natural Language to SQL (NL2SQL) provides a new model-centric paradigm that simplifies database access for non-technical users by converting natural language queries into SQL commands. 
Recent advancements, particularly those integrating Retrieval-Augmented Generation (RAG) and Chain-of-Thought (CoT) reasoning, have made significant strides in enhancing NL2SQL performance. However, challenges such as inaccurate task decomposition and keyword extraction by LLMs remain major bottlenecks, often leading to errors in SQL generation. While existing datasets aim to mitigate these issues by fine-tuning models, they struggle with over-fragmentation of tasks and lack of domain-specific keyword annotations, limiting their effectiveness.
To address these limitations, we present DeKeyNLU, a novel dataset which contains 1,500 meticulously annotated QA pairs aimed at refining task decomposition and enhancing keyword extraction precision for the RAG pipeline. Fine-tuned with DeKeyNLU, we propose DeKeySQL, a RAG-based NL2SQL pipeline that employs three distinct modules for user question understanding, entity retrieval, and generation to improve SQL generation accuracy. We benchmarked multiple model configurations within DeKeySQL RAG pipeline. Experimental results demonstrate that fine-tuning with DeKeyNLU significantly improves SQL generation accuracy on both BIRD (62.31\% to 69.10\%) and Spider (84.2\% to 88.7\%) dev datasets.

\end{abstract}

\section{Introduction}
The rapidly evolving landscape of data accessibility has intensified the need for intuitive interfaces that empower non-technical users to interact with complex databases \cite{javaid2023study, al2024enhancing}. Natural Language to SQL (NL2SQL) systems fulfill this requirement by translating user-friendly natural language commands into precise SQL commands, facilitating seamless information retrieval without requiring users to possess programming skills for database question answering (Database QA) \cite{gao2023text, hong2024next, liu2024survey}.

\begin{figure*}[t]
    \centering
    \includegraphics[width=1\linewidth]{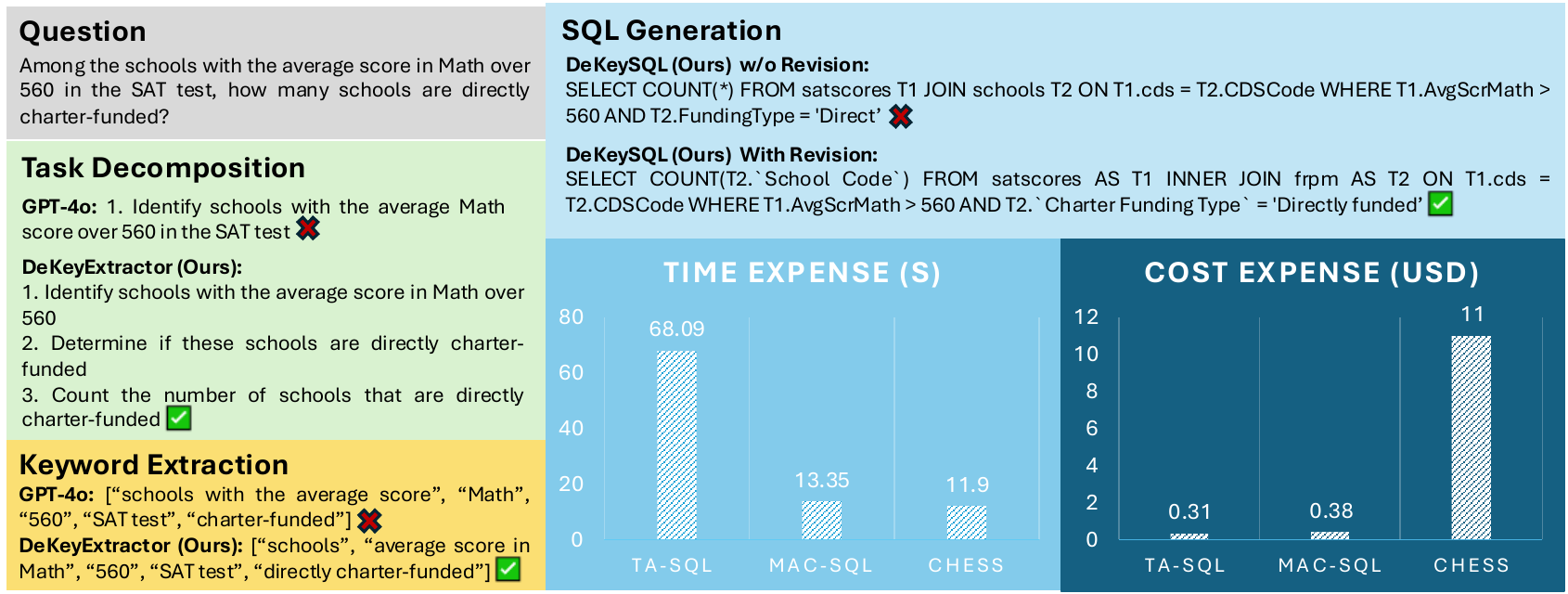}
    \caption{Comparison of advanced NL2SQL methods with DeKeySQL. GPT-4o suffers from incomplete task decomposition and incorrect keyword extraction. Because it lacks a revision module, GPT-4o shows lower code generation accuracy. Methods like MAC-SQL, CHESS, TA-SQL are efficient in either time or cost, but not both.}
    \label{fig:nl2sql_fail_case}
\end{figure*}

Despite considerable advancements in NL2SQL methods, accuracy remains a persistent challenge. Modern hybrid approaches, which integrate Chain of Thought (CoT) \cite{wei2022chain} reasoning and Retrieval-Augmented Generation (RAG) \cite{lewis2020retrieval} with specialized modules—such as  CHASE-SQL \cite{pourreza2024chase}, CHESS \cite{talaei2024chess}, PURPLE \cite{ren2024purple}, DTS-SQL \cite{pourreza2024dts}, and MAC-SQL \cite{wang2023mac}—have made significant strides but continue to encounter two major obstacles: inadequate task decomposition and imprecise keyword extraction from user queries. These issues frequently result in logical errors and incorrect field identifications, particularly when queries involve complex, multi-table relationships.

Prior work in the field has attempted to mitigate these challenges. For instance, QDecomp \cite{tai2023exploring}  and QPL \cite{eyal2023semantic} focus on refining query decomposition techniques by prompting with LLMs, while DARA \cite{fang2024dara} strives to enhance natural language understanding (NLU) through agent frameworks. However, these approaches often lead to over-fragmentation of tasks and do not adequately assess the overall model performance in Database QA contexts. Furthermore, fine-tuning existing models with domain-specific data has shown promise, but prevalent datasets, such as BREAK \cite{wolfson2020break}, lack comprehensive domain-specific annotations for Database QA evaluation and do not emphasize the precise keyword extraction required for database retrieval.

To address these gaps, we present DeKeyNLU, a novel dataset specifically designed to enhance NLU capabilities for NL2SQL systems. DeKeyNLU consists of 1,500 QA pairs meticulously annotated with a focus on two critical aspects: task decomposition and keyword extraction. Originating from the BIRD dataset \cite{li2024can}, it provides a high-quality benchmark for evaluating and improving NL2SQL methods in Database QA.

In addition, we introduce DeKeySQL, a RAG-based pipeline optimized for NL2SQL tasks. DeKeySQL comprises three key modules: (1) User Question Understanding (UQU), which leverages the DeKeyNLU dataset for task decomposition and keyword extraction; (2) Entity Retrieval, incorporating retrieval and re-ranking to identify database elements relevant to the users' question; and (3) Generation, featuring task reasoning and feedback-driven error correction to produce accurate SQL statements.

We benchmarked multiple model configurations within DeKeySQL RAG-based pipeline. Fine-tuning the UQU module with DeKeyNLU improved SQL generation accuracy from 62.31\% to 69.10\% on the BIRD dev dataset and from 84.2\% to 88.7\% on the Spider \cite{yu2018spider} dev dataset. Our experiments reveal that larger models, like GPT-4o-mini\cite{openai20244omini}, excel at task decomposition, while smaller models, such as Mistral-7B \cite{jiang2023mistral}, are more effective for keyword extraction. We also observed that optimal performance varies depending on dataset size and model architecture. Moreover, across the pipeline components, user question understanding emerged as the most significant factor influencing overall SQL generation accuracy, followed by entity retrieval and revision mechanisms. 

\section{Related Work}
\label{sec:Related_Work}

\subsection{Database Question Answering}
Database Question Answering (Database QA) aims to provide precise answers derived from tabular data through advanced reasoning. Early research focused on discrete reasoning \cite{jin2022survey}, with works such as TAT-QA \cite{zhu2021tat}, FinQA \cite{chen2021finqa}, and MVGE \cite{ma2017multi} exploring methods like fine-tuning, pre-training, and in-context learning. Although these approaches advanced the field, they often struggled with generalization in multi-table settings \cite{zhang2024tablellm}.

Parallelly, NL2SQL methods enable mapping natural language questions to SQL commands, offering efficient solutions \cite{gao2023text}. This area spans rule-based, neural network-based, pre-trained language model (PLM)-based, and large language model (LLM)-based strategies \cite{li2024dawn}. Rule-based systems \cite{katsogiannis2021deep} were gradually superseded by neural and transformer-based methods, such as BERT \cite{devlin2018bert}, which improved performance on benchmarks like ScienceBenchmark \cite{zhang2023sciencebenchmark}. Recent advances leverage LLMs (e.g., GPT-4 \cite{achiam2023gpt}), empowering systems such as CHESS \cite{talaei2024chess}, DAIL-SQL \cite{gao2023text}, and MAC-SQL \cite{wang2023mac} with specialized modules for enhanced accuracy and output refinement. Nonetheless, LLM-based approaches continue to face challenges like limited accuracy, high resource costs, and runtime constraints, impeding their practicality \cite{li2024dawn}.

\subsection{Natural Language Understanding for NL2SQL}
Natural Language Understanding (NLU) is central to enabling machines to interpret human language \cite{allen1988natural}, supporting tasks from keyword extraction to complex question answering \cite{yu2023natural,chen-etal-2024-fintextqa}. The application of NLU extends beyond query systems and into different domains \cite{li2025compliance,chen2025climateiqa}. The development of LLMs, including Gemini-Pro \cite{reid2024gemini}, GPT-4 \cite{achiam2023gpt}, and Mistral \cite{jiang2023mistral}, has significantly advanced NLU performance. To further augment their abilities, techniques such as advanced text alignment \cite{zha2024text}, human-provided explanations \cite{liu2021natural}, and explicit reasoning frameworks like Chain-of-Thought (CoT) \cite{wei2022chain} and Tree-of-Thought \cite{yao2024tree} are widely studied. Robustness and generalization are assessed via benchmarks such as Adversarial NLI \cite{nie2019adversarial}, OTTA \cite{deriu2020methodology}, and SemEval-2024 Task 2 \cite{jullien2024semeval}.

In Database QA and NL2SQL, robust NLU is essential for query understanding and decomposition. Methods like QDecomp \cite{tai2023exploring} and QPL \cite{eyal2023semantic} break down complex user questions, while DARA \cite{fang2024dara} and Iterated Decomposition \cite{reppert2023iterated} iteratively refine intent understanding. Grammar-based models (e.g., IRNet \cite{guo2019towards}) and techniques such as ValueNet \cite{brunner2021valuenet} help align natural language with structured schema elements. Yet, accurately handling nuanced queries remains challenging. While datasets like BREAK \cite{wolfson2020break} support research in decomposition, many suffer from over-segmentation, and traditional NLU datasets fail to capture key Database QA aspects like mapping keywords to database elements. Thus, there is a need for more holistic datasets and evaluation protocols that reflect the real-world requirements of Database QA systems.

\section{DeKeyNLU Dataset Creation}
\label{sec:dataset_creation}
As illustrated in Figure \ref{fig:nl2sql_fail_case}, current LLMs exhibit limitations in NLU capabilities for Database QA, which adversely affects NL2SQL accuracy.
Existing datasets like BREAK \cite{wolfson2020break}, while useful, often lack data directly pertinent to the primary task of complex SQL generation.
Moreover, their keyword extraction is often not comprehensive or noise-free enough for robust NLU performance evaluation in Database QA.
To address these challenges, we developed the DeKeyNLU dataset.
\begin{figure*}[t]
    \centering
    \includegraphics[width=1\linewidth]{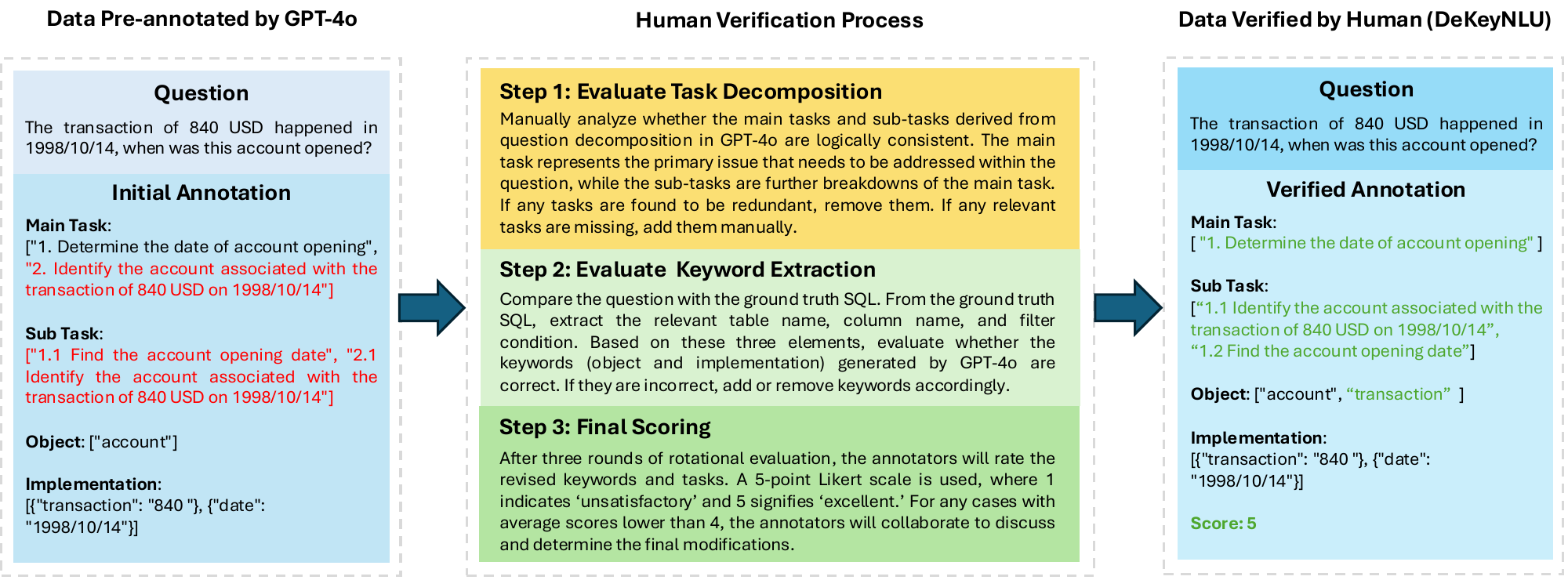} 
    \caption{The DeKeyNLU dataset creation workflow.
User questions are initially pre-annotated by GPT-4o for tasks (main and sub-tasks), objects, and implementations.
These preliminary annotations are then subjected to a rigorous human verification process, where annotators correct and refine both task decomposition and keyword extraction.
This involves three rounds of cross-validation. Following this, a final scoring phase identifies any low-scoring annotations, which are then collaboratively reviewed and further refined to produce the final, high-quality DeKeyNLU dataset.}
    \label{fig:nlu_sample}
\end{figure*}
\subsection{Data Sources}
DeKeyNLU is derived from the BIRD dataset \cite{li2024can}, chosen for its validated origins, large scale, and extensive use in NL2SQL research.
BIRD contains 12,751 text-to-SQL pairs across 95 databases (33.4 GB), spanning 37 professional domains, and is specifically designed for evaluating and training NL2SQL models.
It integrates 80 open-source relational databases from platforms like Kaggle\footnote{\url{https://www.kaggle.com}} and Relation.vit.
An additional 15 relational databases were created for a hidden test set to prevent data leakage.
The BIRD team utilized crowdsourcing to collect natural language questions paired with their corresponding SQL commands.

\subsection{Selection and Annotation}
We randomly selected 1,500 instances from the BIRD training dataset.
Each instance consists of a user question and its ground truth SQL query.
Our annotation process, depicted in Figure \ref{fig:nlu_sample}, began with initial task decomposition and keyword extraction performed by GPT-4o \cite{openai2024fouro}.

Task decomposition involved breaking user questions into a \textit{main task} (primary goal) and \textit{sub-tasks} (refinements of the main task).
Keyword extraction categorized terms into \textit{object} (related to table/column names) and \textit{implementation} (filtering criteria, represented as a dictionary of actions and conditions).
These elements aid similarity matching within the database.

Despite using CoT \cite{wei2022chain} and few-shot techniques \cite{brown2020language}, GPT-4o's initial interpretations were often suboptimal, producing redundant/incomplete tasks or incorrect keywords (see left panel of Figure \ref{fig:nlu_sample}).
This necessitated manual refinement.
Three expert annotators were engaged to review and correct GPT-4o's outputs.
A three-phase cyclic process ensured cross-validation: annotators started with different subsets (A, B, C), then exchanged and reviewed, ensuring each instance was evaluated by all.
The process involved:

\textbf{1. Evaluate Task Decomposition:} Annotators manually assessed the logical consistency of GPT-4o-generated main tasks and sub-tasks, removing redundancies and adding missing relevant tasks.

\textbf{2. Evaluate Keyword Extraction:} Keywords (objects and implementations) were compared against user questions and ground truth SQL elements (filters, table/column names).
Missing keywords were added, and extraneous ones removed.
An initial training on 50 data points helped calibrate annotators and establish quality standards.

\textbf{3.
Final Scoring:} After three rotational evaluation rounds, annotators rated revised keywords and tasks on a 5-point Likert scale (1=unsatisfactory, 5=excellent).
Cases averaging below 4 were discussed collaboratively for final modifications.
The inter-annotator agreement (Krippendorff’s Alpha) for human verification was 0.762, indicating a high level of consistency.

\begin{figure*}[t]
    \centering
    \includegraphics[width=0.8\linewidth]{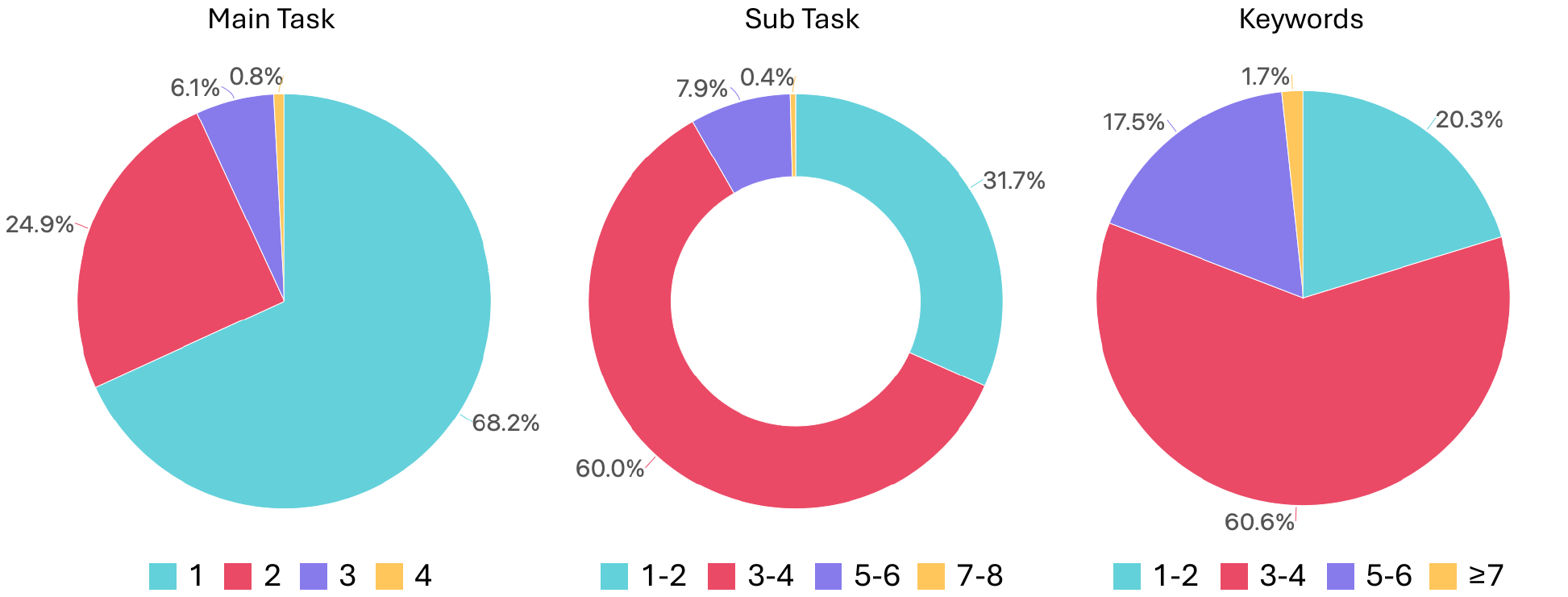} 
    \caption{Distribution of the number of main tasks, sub-tasks, and keywords per question in the DeKeyNLU dataset.
These distributions illustrate the complexity inherent in the questions, reflecting the reasoning and integration capabilities required of NL2SQL models.}
    \label{fig:nl2sql_data_statistics}
\end{figure*}

\subsection{Dataset Statistics}
After three review rounds, a dataset of 1,500 question-answer pairs (decomposed tasks and keywords) was finalized.
It was partitioned into training (70\%), validation (20\%), and testing (10\%) sets for robust model development and evaluation.
Figure \ref{fig:nl2sql_data_statistics} shows the distribution of main tasks, sub-tasks, and keywords, which indicate question complexity.
A higher number of tasks tests reasoning and integration capabilities, while more keywords suggest intricate table/column setups prone to errors.
For main tasks: 68.2\% of questions have one task, 24.9\% have two, and 6.9\% have three or more.
For sub-tasks: 31.7\% comprise one to two sub-tasks, 60\% have three to four, and 8.3\% contain over five.
For keywords: 20.3\% are linked to one or two keywords, 60.6\% to three or four, and 19.2\% to five or more.

\section{RAG-based NL2SQL Framework: DeKeySQL}
\begin{figure*}[t]
    \centering
    \includegraphics[width=1\linewidth]{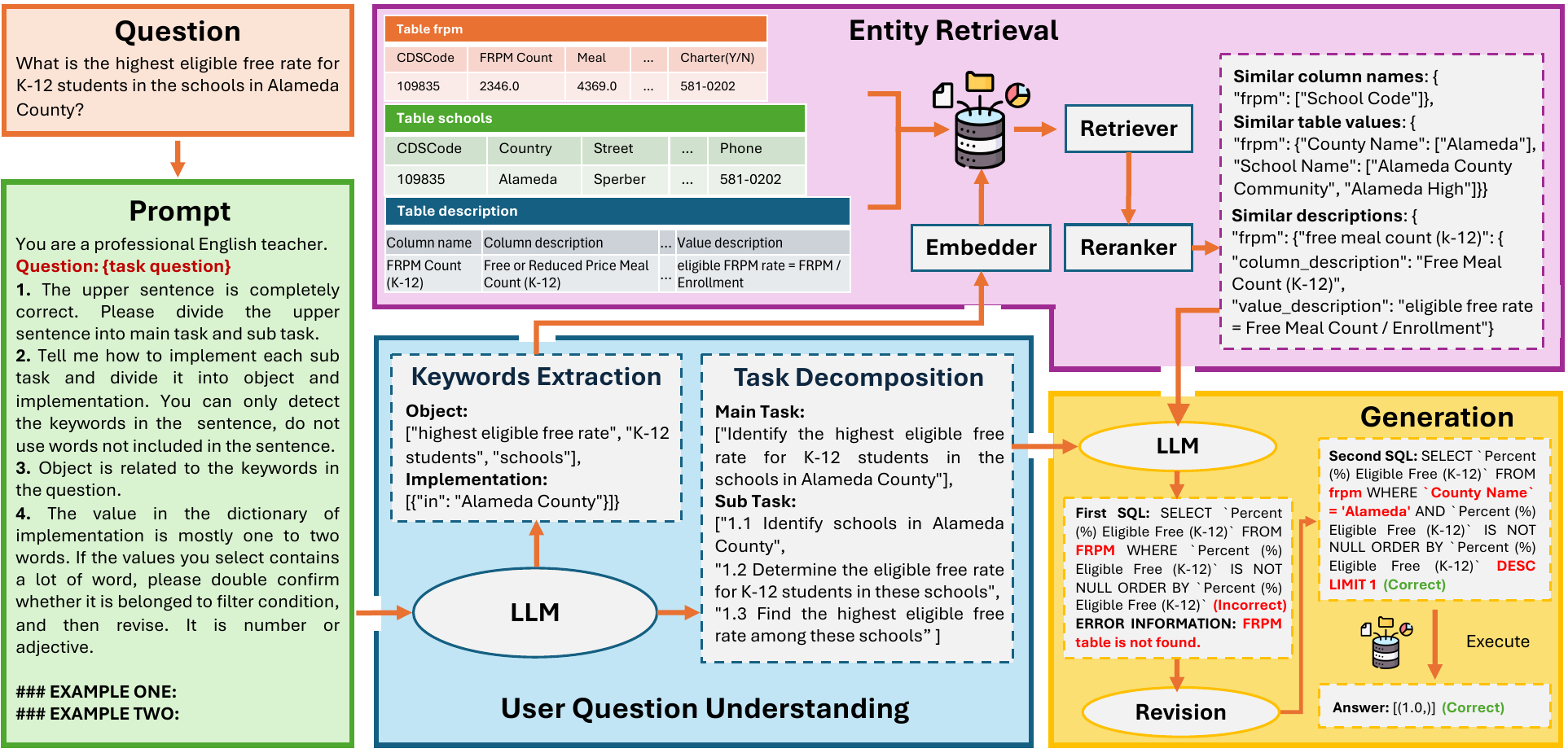} 
    \caption{The DeKeySQL Framework.
(1) The user's question is processed by the User Question Understanding (UQU) module using a prompt template, directing an LLM (fine-tuned on DeKeyNLU) to perform keyword extraction and task decomposition.
(2) Extracted keywords are fed to the Entity Retrieval module to identify relevant column names, table values, and descriptions from the database.
(3) Task decomposition outputs, retrieved entity data, and the original question are then input to the Generation LLM to produce SQL code.
(4) If errors occur, the error information and generated SQL are passed to a revision LLM for correction.
(5) Finally, the corrected SQL is executed to obtain the answer.}
    \label{fig:nl2sql-framework}
\end{figure*}

We introduce DeKeySQL, a novel RAG-based framework for NL2SQL generation, designed to address common issues in existing approaches like MAC-SQL \cite{wang2023mac} and CHESS \cite{talaei2024chess}, such as long runtimes, high costs, and accuracy limitations.
As depicted in Figure \ref{fig:nl2sql-framework}, DeKeySQL comprises three main components: User Question Understanding (UQU), Entity Retrieval, and Generation.

\subsection{User Question Understanding (UQU)}
The initial phase of DeKeySQL focuses on comprehending user questions (Figure \ref{fig:nl2sql-framework}).
The user question is incorporated into a prompt template and fed to an LLM (fine-tuned on DeKeyNLU) to generate a structured response encompassing two key tasks: Task Decomposition and Keyword Extraction.

\noindent\textbf{Task Decomposition:} Inspired by CoT reasoning \cite{wei2022chain}, we decompose complex user questions into manageable components.
We employ a two-step CoT approach, breaking questions into a \textit{main task} (primary goal) and \textit{sub-tasks} (refinements).
This hierarchical structure aids the generation model by clarifying task dependencies.
For example, the main task often corresponds to the main SELECT component in SQL, while sub-tasks map to operations like INNER JOIN, WHERE, etc. General LLMs like GPT-4o can falter here (Figure \ref{fig:nlu_sample});
thus, supervised fine-tuning with DeKeyNLU is employed to enhance stability and reliability.

\noindent\textbf{Keyword Extraction:} Unlike prior methods that simply broke sentences into keywords leading to irrelevancy, we classify keywords into \textit{object} (terms associated with table/column names) and \textit{implementation} (filtering criteria as a key-value dictionary).
While In-Context Learning (ICL) with multiple examples can guide LLMs, models like GPT-4o may still generate irrelevant keywords (Figure \ref{fig:nlu_sample}).
To mitigate this, we fine-tune smaller models like Mistral-7B \cite{jiang2023mistral} using DeKeyNLU, enhancing keyword extraction accuracy.

\subsection{Entity Retrieval}
Following keyword extraction, this module retrieves corresponding database entities: table names, column names, table values, and textual descriptions (column/value descriptions).
It consists of an embedder, retriever, and re-ranker.
All table data are initially encoded and stored in a Chroma database.
Keywords from UQU are encoded by the embedder and then used by the retriever to find the top-five resembling entities from the database.
These are then passed to a re-ranker, which recalculates similarity scores and selects the two most similar entities.
This process is divided into two sub-tasks:

\noindent\textbf{Database Retrieval:} Retrieves column names and table values.
To handle large volumes of database values efficiently, we use MinHash \cite{zhu2016lsh} + Jaccard Score or BM25 \cite{robertson2009probabilistic}.
MinHash generates fixed-size signatures for sets, approximating Jaccard similarity, while BM25 is a probabilistic model using term frequency and inverse document frequency.
For column names, the top five scoring entities (score > 0) are selected.
For purely numeric keywords, only exact matches for table values are considered;
for mixed text/numeric keywords, the top five scoring entities are selected without a threshold.
These are re-ranked to find the two most similar entities.
Retrieved entities are cross-referenced to get table/column names, then de-duplicated and categorized (Figure \ref{fig:nl2sql-framework}).

\noindent\textbf{Textual Description Retrieval:} Retrieves column and value descriptions. Given a smaller dataset for this task, we directly use an embedding model to encode data, then cosine similarity in the retriever to find the top five entities.
A specialized re-ranker model then determines the final relevance order.

\subsection{Generation}
This process has two phases: SQL Generation and Revision.

\noindent\textbf{SQL Generation:} Using ICL, general LLMs like GPT-4o \cite{openai2024fouro} generate initial SQL statements.
Prompts (Appendix Figure \ref{fig:candidate_generation}) are structured into: data schema (formats, names, examples from Entity Retrieval), user question reasoning (question, main/sub-tasks from UQU, hints from dataset), constraints, and incentives.
These details guide the model to produce an initial SQL statement.

\noindent\textbf{Revision:} Initial SQL may contain errors (incorrect table names, misaligned columns, etc., see Figure \ref{fig:nl2sql-framework}).
Erroneous SQL and corresponding error messages are fed back to an LLM for revision (Appendix Figure \ref{fig:revision}).
This iterative process yields syntactically correct, operational SQL commands.

\section{Experiments}
We conducted comprehensive experiments to evaluate the DeKeyNLU dataset and the DeKeySQL system.
Our aims were: (1) to demonstrate DeKeyNLU's effectiveness for model fine-tuning, and (2) to assess DeKeySQL's performance against open-source SOTA methods.

\subsection{Experiment Setting}

\noindent\textbf{Datasets:} We used three datasets: DeKeyNLU (our proposed dataset), the BIRD development dataset, and the Spider dataset.

\noindent\textbf{NL2SQL Baseline Selection:} We selected open-source NL2SQL methods or those with published papers, including GPT-4o as a baseline.
Methods include: Distillery \cite{maamari2024death} (schema linking augmentation), CHESS \cite{talaei2024chess} (integrates data catalogs/values), MAC-SQL \cite{wang2023mac} (multi-agent framework), Dail-SQL \cite{gao2023text} (prompt engineering with advanced question/example handling), and CodeS-15B \cite{li2024codes} (incremental pre-training on SQL-centric corpus).

\noindent\textbf{Base Model Selection:} For UQU, models included GPT-4o-mini \cite{openai20244omini}, GPT-4 \cite{achiam2023gpt}, Mistral-7B \cite{jiang2023mistral}, LLaMA3-8B \cite{dubey2024llama}, Baichuan2-7B, and 13B \cite{yang2023baichuan}.
For entity retrieval, MinHash \cite{zhu2016lsh} + Jaccard Score was compared against BM25 \cite{robertson2009probabilistic}.
Embedding models assessed: text-embedding-3-large \cite{openai2024embedding}, Stella-1.5B, and Stella-400M.
For code generation fine-tuning: DeepSeek-Coder-V2-Instruct, DeepSeek-Coder-V2-Base \cite{zhu2024deepseek}, and Qwen-1.5-Coder \cite{yang2024qwen2}.

\noindent\textbf{Fine-tuning Process:} Conducted on 4 Nvidia 4090 GPUs using Distributed Data Parallel and DeepSpeed.
Uniform batch size of 1, epoch count of 1, learning rate of 2e-4.
Low-Rank Adaptation (LoRA) \cite{hu2021lora} was used with rank=64, alpha=16, dropout=0.05.
Bit precision was 4. Fine-tuning a UQU model with DeKeyNLU took ~30 minutes;
a code generation model took 4-5 hours.
\begin{table}[t]
\small
\centering

\resizebox{0.48\textwidth}{!}{
\begin{tabular}{lcccc}
\toprule
\multirow{2}{*}{\bf Method} & \multicolumn{3}{c}{\bf Task Decomposition} & \multicolumn{1}{c}{\bf Keyword Extraction}\\ 
\cmidrule(lr){2-4} \cmidrule(lr){5-5}
&{\bf BLEU} & {\bf ROUGE} & {\bf GPT-4o} & {\bf F1 Score } \\ \midrule
LLaMA-8B&0.679 &0.813 &4.141 &0.677     \\
Baichuan2-7B& 0.616 &0.697 &4.112 &0.511 \\
Baichuan2-13B&0.622 &0.722 &4.124 &0.583 \\
Mistral-7B& 0.706 &{0.798} &4.081 &{\bf 0.696}\\
GPT-4o-mini&0.713 &0.811 &4.256 &{ 0.672}\\
GPT-4&{\bf 0.722} &{\bf 0.816} &{\bf 4.286} &0.665 \\ \bottomrule
\end{tabular}}
\caption{Performance comparison of various fine-tuned models on task decomposition (BLEU, ROUGE, GPT-4o score) and keyword extraction (F1 Score) using the DeKeyNLU test set.
GPT-4 leads in task decomposition metrics, while Mistral-7B shows the best F1 score for keyword extraction.}
\label{tab:finetuned_model_performance} 
\label{finetuned_model} 
\end{table}

\begin{table}[t]
\small
\centering

\resizebox{0.48\textwidth}{!}{
\begin{tabular}{lc}
\toprule 
\multicolumn{1}{c}{\bf Method Configuration}  &\multicolumn{1}{c}{\bf Dev EX} \\ \midrule
UQU + Entity Retrieval + Revision + Generation &60.36   \\ 
Entity Retrieval + Revision + Generation & 55.28 \\
Entity Retrieval + Generation & 51.25 \\
Generation Only         &46.35   \\ \bottomrule
\end{tabular} }
\caption{Module ablation study for DeKeySQL with GPT-4 as 
the backbone on the BIRD development set, showing Dev EX improvement with each added component.
The full pipeline (UQU + Entity Retrieval + Revision + Generation) achieves the highest accuracy.}
\label{tab:module_ablation_study} 
\label{module_ablation} 
\end{table}

\subsection{Metrics}
\noindent\textbf{BLEU, ROUGE, and GPT-4o Score:} For evaluating task decomposition in NLU, we compared generated reasoning results against human-labeled ground truth using BLEU (BLEU-1, BLEU-2 for linguistic accuracy via n-gram matches) \cite{papineni2002bleu}, ROUGE (ROUGE-1, ROUGE-2, ROUGE-L for n-gram, sequence, and word pair overlap, measuring comprehensiveness/relevance) \cite{lin2004rouge}, and GPT-4o scores (five-point Likert scale, calibrated with human judgment 
for overall similarity) \cite{zheng2023judging}. Calibration details are in Table \ref{tab:calibration}.

\noindent\textbf{F1 Score:} For keyword extraction in NLU, performance was evaluated using precision, recall, and the F1 score, balancing correctness and recall for a holistic view of extraction efficiency.

\noindent\textbf{Execution Accuracy (EX):} SQL query correctness was measured by comparing executed predicted query results against reference query results on specific database instances.
This ensures semantic correctness and accounts for varied SQL formulations yielding identical results.

\begin{table*}[t]
\centering
\small

\resizebox{0.8\textwidth}{!}{
\begin{tabular}{l|ccccc|ccc|c}
\toprule
\multicolumn{1}{c|}{\multirow{2}{*}{Method}}& \multicolumn{5}{c|}{Dataset Size}   & \multicolumn{3}{c|}{Epoch} & \multicolumn{1}{c}{\multirow{2}{*}{\begin{tabular}[c]{@{}c@{}} w/o \\Fine-tuning \end{tabular}}} \\ \cline{2-9} 
& 20\% & 40\% & 60\% & 80\% & 100\%  & 1  & 2  & 3 &  \\ \midrule
\multicolumn{1}{l|}{LLaMA-8B}&0.609 &0.636 &{\bf 0.677} &0.661 &0.653 &0.677 &0.728 &0.734 &0.442   \\
\multicolumn{1}{l|}{Baichuan2-7B}&0.497 
&0.515 &0.558 &0.522 &0.511 &0.511 &0.648 &0.688 &0.208 \\
\multicolumn{1}{l|}{Mistral-7B}&{\bf 0.648} &{\bf 0.640 }&0.634 &{\bf 0.694} &{\bf 0.696} &{\bf 0.696} &{\bf 0.755} &{\bf 0.769} &{\bf 0.502} \\
\multicolumn{1}{l|}{Baichuan2-13B}&0.412 &0.554 &0.573 &0.638 &0.585 
&0.585 &0.609 &0.647 &0.266 \\ \bottomrule
\end{tabular}}
\caption{Impact of DeKeyNLU dataset size (percentage of training data used) and number of training epochs on keyword extraction F1 score for various models.
Mistral-7B generally benefits from more data and epochs, while other models show nuanced responses to dataset size.}
\label{tab:dataset_size_epochs_impact} 
\label{dataset_size} 
\end{table*}

\subsection{Results}
\noindent\textbf{Supervised Fine-tuning with DeKeyNLU:}
As shown in Table \ref{model_ablation} (top row vs third row for UQU impact), fine-tuning the UQU module with DeKeyNLU elevated Dev EX from 62.31\% (GPT-4o without DeKeyNLU fine-tuning for UQU) to 69.10\% (GPT-4o-mini fine-tuned on DeKeyNLU for UQU, with GPT-4o for generation) on the BIRD dev dataset.
On the Spider dev dataset, a similar improvement from 84.2\% to 88.7\% was observed.
Table \ref{finetuned_model} reveals that model size impacts suitability for different fine-tuning tasks.
Larger models (GPT-4, GPT-4o-mini) perform better on complex tasks like task decomposition after fine-tuning.
Smaller models (Mistral-7B) outperform on tasks not requiring deep understanding, like keyword extraction.
This suggests task-specific model selection for fine-tuning is crucial.
Table \ref{dataset_size} shows the effects of varying dataset sizes and epochs on keyword extraction fine-tuning.
Mistral-7B performed best overall, followed by LLaMA-8B. For all models except Mistral-7B, F1-Score initially rose then fell with increasing training data, indicating more data is not always better.
Increasing epochs consistently improved F1-Scores, suggesting it's a highly effective method for enhancing keyword extraction accuracy.

\begin{table*}[t]
\centering

\resizebox{0.8\textwidth}{!}{
\begin{tabular}{llllc}
\toprule
\textbf{Module Focus} & \multicolumn{1}{l}{\textbf{UQU Model}} & \multicolumn{1}{l}{\textbf{Entity Retrieval (Retriever + Embedder)}} & \multicolumn{1}{l}{\textbf{Generation Model}} & \multicolumn{1}{c}{\textbf{Dev EX}} \\ \midrule
\multirow{4}{*}{\textbf{UQU}} 
    & GPT-4o-mini (Finetuned on DeKeyNLU) & MinHash + Stella-400M & GPT-4o & \textbf{69.10} \\
    & Mistral-7B (Finetuned on DeKeyNLU) & MinHash + Stella-400M & GPT-4o & 65.16 \\
    & GPT-4o (No DeKeyNLU fine-tuning) & MinHash 
+ Stella-400M & GPT-4o & 62.31 \\ 
    
 & GPT-4 (No DeKeyNLU fine-tuning) & MinHash + Stella-400M & GPT-4o & 59.62 \\ \midrule
\multirow{5}{*}{\textbf{Generation}}
    & GPT-4 & MinHash + Stella-400M & GPT-4o & 59.62 \\
    & GPT-4 & MinHash + Stella-400M & DeepSeek-Coder-V2-Instruct (Finetuned) & 55.78 \\
    & GPT-4 & MinHash + Stella-400M & DeepSeek-Coder-V2-Base (Finetuned) & 50.41 \\
    & GPT-4 & MinHash + Stella-400M & Qwen-1.5-Coder (Finetuned) & 30.82 \\
    & GPT-4 & MinHash + Stella-400M & GPT-4 & 53.17 \\ \midrule
\multirow{4}{*}{\textbf{Entity Retrieval}}
   
  & GPT-4 & MinHash + 
 Stella-400M & GPT-4 & 53.17 \\
    & GPT-4 & MinHash + Stella-1.5B & GPT-4 & 51.36 \\
    & GPT-4 & MinHash + text-embedding-3-large & GPT-4 & 51.25 \\
    & GPT-4 & BM25 + text-embedding-3-large & GPT-4 & 49.34 \\ \bottomrule
\end{tabular}}
\caption{Performance (Dev EX on BIRD) of DeKeySQL with different backbone models for UQU, Entity Retrieval, and Generation modules.
Results highlight the impact of DeKeyNLU fine-tuning (e.g., GPT-4o-mini for UQU) and the surprising efficacy of smaller embedding models like Stella-400M.}
\label{tab:model_backbone_ablation} 
\label{model_ablation} 
\end{table*} 

\noindent\textbf{Ablation Study:}
The module ablation study (Table \ref{module_ablation}), using GPT-4 as the backbone, showed significant accuracy improvements with each added module.
The UQU module (keyword extraction and task decomposition) yielded the largest gain, boosting accuracy by 9.18\% (from 51.25\% to 60.36\%, comparing "Entity Retrieval + Generation" with the full pipeline).
The entity retrieval module also contributed substantially, increasing accuracy by 4.9\% (from 46.35\% to 51.25\%, comparing "Generation" with "Entity Retrieval + Generation").
UQU, entity retrieval, and revision modules were all indispensable.
The model ablation study (Table \ref{model_ablation}) indicated MinHash outperformed BM25 in entity retrieval (e.g. 51.25\% vs 49.34\% when other components are GPT-4 and text-embedding-3-large) with less computation time.
Surprisingly, the smaller Stella-400M embedding model surpassed the larger Stella-1.5B (e.g., 53.17\% vs 51.36\% Dev EX with GPT-4 UQU/Gen and MinHash retriever), suggesting parameter size isn"t always a guarantor of better performance.
For code generation, general LLMs like GPT-4o and GPT-4 outperformed the fine-tuned smaller code models in our setup, underscoring the impact of parameter size and pre-training quality on complex code generation accuracy.
These findings emphasize balancing architecture, parameter size, and task-specific optimization.

\noindent\textbf{BIRD and Spider Dataset Evaluation:}
For BIRD, we report Dev EX due to its anonymity policy for test set evaluation;
test EX and VES will be added in future updates.
Table \ref{BIRD-Result} shows DeKeySQL achieves the best Dev EX on BIRD compared to other SOTA models and is the current best open-source method.
On Spider, DeKeySQL shows the highest EX on both dev and test sets.
In practical utility assessment (Table \ref{practical_utility}), DeKeySQL excels in time efficiency, operational cost, and accuracy.
Compared to CHESS, DeKeySQL achieves a 52.4\% runtime reduction and a 97\% operational cost decrease, showcasing significant industrial application potential.
DeKeySQL is the top-performing open-source NL2SQL method evaluated.

\begin{table}[t]
\centering
\small

\resizebox{0.45 \textwidth}{!}{
    \begin{tabular}{l|c|cc}
        \toprule 
        \multicolumn{1}{c|}{\multirow{2}{*}{\bf Method}} & \multicolumn{1}{c|}{\bf BIRD Dataset}   & \multicolumn{2}{c}{\bf Spider Dataset}  \\ \cline{2-4} 
        & \multicolumn{1}{c|}{\bf Dev EX} 
        & \multicolumn{1}{c}{\bf Dev EX} & \multicolumn{1}{c}{\bf Test EX}  \\ \midrule
        GPT-4         & 46.35  & 74.0 & 
 67.4  \\
        Distillery \cite{maamari2024death}         & 67.21  & - & -  \\ 
        CHESS \cite{talaei2024chess}            & 68.31  & 87.2 & - \\ 
        Dail-SQL \cite{gao2023text}             & 54.76  & 84.4 & 86.6  \\ 
     
    SFT CodeS-15B \cite{li2024codes}        & 58.47  & 84.9 & 79.4  \\ 
        MAC-SQL \cite{wang2023mac}          & 57.56  & 86.7 & 82.8  \\ 
        \bf DeKeySQL(ours) & \bf 69.10 & \bf 88.7 & \bf 87.1  \\ \bottomrule
    \end{tabular}
}
\caption{Performance comparison (Execution Accuracy - EX) on BIRD and Spider datasets.
DeKeySQL demonstrates state-of-the-art results among evaluated methods, particularly on the development sets. "-" indicates results not reported or not applicable.
Results are sourced from official leaderboards where available.}
\label{tab:bird_spider_results} 
\label{BIRD-Result} 
\end{table}

\begin{table}[t]
\centering

\resizebox{0.45 \textwidth}{!}{
    \begin{tabular}{lccc}
        \toprule 
        \multicolumn{1}{c}{\bf Method} & \multicolumn{1}{c}{\bf Time(s)} & \multicolumn{1}{c}{\bf Dev EX} & \multicolumn{1}{c}{\bf Cost (USD)} \\ \midrule 
        CHESS \cite{talaei2024chess}            & 119.38 & 0.5 & 11 \\ 
       
  TA-SQL             & 57.92 & 0.5 & 0.41 \\
        SFT CodeS-15B \cite{li2024codes}        & \bf 35 & 0.4 & - \\ 
        MAC-SQL \cite{wang2023mac}          & 133.55 & 0.7 & 0.38 \\ 
        Chat2Query         & 680.96 & 0.6 & - \\ 
        \bf DeKeySQL (ours)  &  56.81 & \bf 0.8 & \bf 0.32 \\ \bottomrule 
    \end{tabular}
}
\caption{Practical utility metrics (Time, Dev EX, Cost) for NL2SQL methods using GPT-4o as the base generation model.
DeKeySQL demonstrates a strong balance of efficiency and accuracy. "-" indicates data not available.}
\label{tab:practical_utility_metrics} 
\label{practical_utility} 
\end{table}

\section{Conclusion}
This paper introduced DeKeyNLU, a novel dataset of 1,500 annotated question-SQL pairs, specifically designed to tackle critical challenges in task decomposition and keyword extraction for NL2SQL systems.
Built upon the BIRD dataset, DeKeyNLU furnishes domain-specific annotations and establishes a high-quality benchmark for evaluating and improving LLM performance in this domain.
Our comprehensive experiments demonstrate that fine-tuning with DeKeyNLU significantly enhances SQL generation accuracy, with performance reaching 69.10\% on the BIRD development set (an increase from 62.31\%) and 88.7\% on the Spider development set (up from 84.2\%). We further observed that larger models like GPT-4o-mini are particularly adept at task decomposition, whereas smaller, more agile models such as Mistral-7B excel in keyword extraction.
Within the NL2SQL pipeline, entity retrieval was identified as the most critical component for overall accuracy, followed by user question understanding and the revision mechanisms.
These findings underscore the profound value of dataset-centric approaches and meticulous pipeline design in advancing the capabilities of NL2SQL systems, paving the way for intuitive and accurate data interaction for users.

\section*{Limitations}
While DeKeyNLU and DeKeySQL demonstrate considerable advancements, several limitations and avenues for future research remain.
The primary constraint is the DeKeyNLU dataset size (1,500 samples), a consequence of resource limitations.
While meticulously curated, this size may affect the robustness and generalizability of UQU models, particularly for highly diverse real-world scenarios.
Expanding this dataset, possibly through semi-automated annotation techniques or exploring data augmentation strategies tailored for structured NLU tasks, is a key future direction.
The restricted availability of high-quality annotated data, often compounded by copyright concerns for source data, poses an ongoing challenge for dataset expansion and community sharing that the field must address.
Our benchmarking was also constrained by computational resources, preventing experimentation with the largest available LLMs (e.g., DeepSeek-V2-Coder-236B, Llama3.1-70B) or their integration with more advanced RAG modules.
Such larger models and components could potentially yield further accuracy and robustness improvements.
Future work should evaluate these cutting-edge models and diverse RAG configurations to establish more comprehensive benchmarks.
Additionally, exploring the generalization of DeKeyNLU-trained models to completely unseen database schemas and question types, beyond the scope of BIRD, would be valuable.
Investigating adaptive task decomposition strategies that can dynamically adjust granularity based on query complexity is another promising research avenue.

\section{Acknowledgements}


This work is supported by the Guangzhou-HKUST(GZ) Joint Funding Program (No. 2024A03J0630) and NSFC (No. 62402413). Additional funding was kindly provided by HSBC.

\bibliography{custom} 
\appendix

\section{Disclaimers}
DeKeyNLU is developed based on the BIRD dataset. Given the BIRD dataset's claim that it should be distributed under \textbf{CC BY-NC 4.0} \cite{li2024can}.

The DeKeyNLU dataset will be publicly available under the same \textbf{CC BY-NC 4.0} license.

All annotators in our team hold bachelor's degrees with their education conducted in English.
They were compensated at a rate of 14 USD/hr for their annotation work.
The DeKeyNLU dataset does not contain any personally identifiable information or offensive content.

The DeKeyNLU dataset was initially generated by an LLM and then meticulously annotated and revised by human experts.
After three rounds of manual revision, the high-quality DeKeyNLU dataset was finalized.

\section{Accessibility}

1. Links to access the dataset and its metadata. (\url{https://huggingface.co/datasets/GPS-Lab/DeKeyNLU}). Link to access code (\url{https://github.com/AlexJJJChen/DeKeyNLU}).

2. The data is saved in both json and csv format, where an example is shown in the README.md file.

3. Logos AI Lab research group will maintain this dataset on the official GitHub account.

4. CC-BY-NC-4.0 (\url{https://github.com/AlexJJJChen/DeKeyNLU/blob/main/LICENSE}).
\section{Performance of Revision Module}
In our analysis of DeKeySQL, the Revision module is activated only once during processing.
While it enhances accuracy, multiple iterations do not necessarily lead to better outcomes proportionally to cost.
We experimented on 50 sample queries, varying the revision threshold from 1 to 5 (Table \ref{tab:revision_performance_thresholds}).
Increasing the threshold generally improves accuracy with an associated rise in computational cost, though execution time doesn"t follow a consistent pattern.
A threshold of 3 offered an optimal balance of cost, accuracy, and execution time for our setup.
For instance, a revision threshold of 1 yielded a Dev EX of 67.28\% (on the BIRD dev set, based on context of DeKeySQL performance improvements being on BIRD), while increasing it to 5 improved Dev EX to 69.10\%.
The Revision module is capped at a threshold of 5 to prevent infinite loops and manage cost-effectiveness.
\begin{table}[h]
    \centering
    \small

    \begin{tabular}{cccc}
    \toprule
\textbf{Threshold}	  &	  \textbf{Time (s)}	  &	  \textbf{Cost (USD)}	  &	  \textbf{Accuracy (\%)}	  \\ \midrule
One	  &	  322.79	  &	  1.402	  &	  48	  \\ 
Two	  &	  357.57	  &	  1.598	  &	  58	  \\
Three	  &	  339.44	  &	  2.953	  &	  62	  \\
Four	  &	  345.23	  &	  3.119	  &	  62	  \\
Five	  &	  469.04	  &	  4.265	  &	  64    \\ \bottomrule
    \end{tabular}
\caption{Performance of the Revision module with different iteration thresholds on a sample of 50 queries from the BIRD dev set.
Accuracy refers to Dev EX.}
    \label{tab:revision_performance_thresholds}
    \label{tab:revision} 
\end{table}

\section{Error Analysis Details}
\label{sec:Error_Analysis_Details}
Previous research, such as CHESS and CHASE-SQL, has not disclosed the specific datasets used for their error analyses, making direct objective comparisons challenging.
Therefore, we conducted our own error analysis by randomly sampling 20\% of failed cases from the BIRD dataset results.
As shown in Table \ref{error_rario}, we found that 45\% of the golden SQL commands themselves had issues, primarily incorrect column names (11\%) and missing GROUP BY/DISTINCT/RANK clauses (8\%).
Additionally, 6\% of golden SQLs did not follow provided evidence cues.
For DeKeySQL, 49\% of its incorrectly generated SQL commands were mainly due to not adhering to evidence (17\%), incorrect column usage (11\%), and incorrect operations (8\%).
Vague questions, where question information was insufficient for correct SQL generation, accounted for 6\% of issues, affecting both golden and predicted SQL.

\begin{table}[t]
\centering

\resizebox{0.45\textwidth}{!}{
\begin{tabular}{lccc} \toprule 
\textbf{Error Category}	  &	  \textbf{\% in Incorrect} & \textbf{\% in Incorrect} &	 \textbf{\% in Vague} \\
& \textbf{Golden SQL (Total 45\%)} & \textbf{Predicted SQL (Total 49\%)} & \textbf{Questions (Total 6\%)} \\ \midrule
Evidence Misalignment	  &	  6\%	  &	  17\%	  &	  0\%	  \\
Incorrect Column	  &	  11\%	  &	  11\%	  &	  5\%	  \\
Incorrect Filtering	  &	  
 5\%	  &	  4\%	  &	  1\%	  \\
Description Issue	  &	  0\%	  &	  0\%	  &	  0\%	  \\
Incorrect Aggregation	  &	  2\%	  &	  1\%	  &	  0\%	  \\
Group by/Distinct/Rank Issue &	  8\%	  &	  6\%	  &	  0\%	  \\
Incorrect Operation	  &	  6\%	  &	  8\%	  &	  0\%	  \\
Date Handling Error	  &	  0\%	  &	  0\%	  &	  0\%	  \\
NULL Value Handling	  &	  3\%	  &	  1\%	  &	  
 0\%	  \\
Revision Error (Internal) &	  0\%	  &	  0\%	  &	  0\%	  \\
Incorrect Table	  &	  4\%	  &	  1\%	  &	  0\%	  \\ \bottomrule 
    \end{tabular}}
\caption{Distribution of error categories identified in Golden SQL commands, DeKeySQL's Predicted SQL commands (for failed cases), and Vague Questions from a 20\% sample of BIRD dataset failed cases.}
\label{tab:error_distribution_analysis}
    \label{error_rario} 
\end{table}

\subsection{Error Types in Predicted SQL of DeKeySQL}
Our error analysis (examples in Tables \ref{tab:column} through \ref{tab:Filtering}) identified five significant error types in DeKeySQL's predicted SQL:
\begin{itemize}
    \item \textbf{Incorrect Column Names:} DeKeySQL sometimes generates inaccurate column names or selects incorrect columns.
    \item \textbf{Incorrect Aggregation:} Occasionally, DeKeySQL joins tables unnecessarily or uses incorrect column names in the "ON" clause, leading to aggregation issues.
    \item \textbf{Incorrect Operation:} DeKeySQL may exhibit flawed or superfluous mathematical calculations, often due to an insufficient understanding of the database schema.
    \item \textbf{Incorrect Evidence Understanding:} In some instances, DeKeySQL fails to consult relevant evidence (e.g., formulas provided in prompts) when generating SQL commands, highlighting limitations in the LLM's adherence to complex instructions.
    \item \textbf{Incorrect Filtering:} Filtering values in SQL commands can be inaccurate or non-existent in the database.
This is often linked to imprecise keyword extraction, indicating room for improvement in that sub-module.
\end{itemize}
\subsection{Errors in Keyword Extraction and Task Decomposition (Qualitative)}
Qualitative examples of errors in keyword extraction and task decomposition from the DeKeyNLU annotation process are presented in Table \ref{tab:keyword_extraction_errors} and \ref{tab:task_decomposition_errors}.
Keyword extraction errors are categorized as: missed keywords, wrong keywords, and useless keywords.
Task decomposition errors include: a main task that should be a sub-task, an incomplete main task, an incomplete sub-task, incorrect sub-task numbering/assignment, and ambiguous sub-task definitions.
These examples informed the refinement of our annotation guidelines and highlight the challenges LLMs face.
\section{Calibration of GPT-4o Score for NLU Evaluation}
To confirm the reliability of GPT-4o's automated evaluation for NLU tasks (task decomposition) and measure its alignment with human judgments, we incorporated calibration techniques.
We compared GPT-4o scores for generated answers with human scores assigned by the three dataset annotators on a 5-point Likert scale.
Table \ref{tab:calibration} summarizes results for six models.
On average, human evaluations were consistently slightly higher (0.125 to 0.202, mean 0.152) than GPT-4o scores.
To address this, inspired by methodologies from EvalGen \cite{shankar2024validates} and CalibraEval \cite{li2024calibraeval}, we performed a calibration between GPT-4o scores and human evaluations.
The resulting regression model is:
\begin{equation}
\text{HumanEvaluation} = 1.015 \times \text{GPT4oScore} + 0.042
\label{eq:calibration}
\end{equation}
After applying this calibration (Equation \ref{eq:calibration}), the average difference between calibrated GPT-4o scores and human evaluations reduced from 0.152 to 0.046, demonstrating significantly improved consistency.

\begin{table*}[h]
    \centering

    \resizebox{0.9\textwidth}{!}{ 
    \begin{tabular}{lccccc}
        \toprule
        \textbf{Model} & \textbf{GPT-4o Score} & \textbf{Human Evaluation} & \textbf{Difference (Human - GPT-4o)} & \textbf{Calibrated GPT-4o Score} & \textbf{Difference (Human - Calibrated)} \\ \midrule
        LLaMA-8B & 4.141 & 4.266 & 0.125 & 4.245 & 0.021 \\ 
        Baichuan2-7B & 4.112 & 4.25 & 0.138 
 & 4.216 & 0.034 \\ 
        Baichuan2-13B & 4.124 
 & 4.316 & 0.192 & 4.228 & 0.088 \\ 
        Mistral-7B & 4.081 & 4.283 & 0.202 & 4.184 & 0.099 \\ 
        GPT-4o-mini & 4.256 & 4.383 & 0.127 & 4.362 & 0.021 \\ 
        GPT-4 & 4.286 & \textbf{4.416} & 0.13 & \textbf{4.392} & 0.024 \\ \bottomrule
    \end{tabular}}
    \caption{Comparison of Model Scores (GPT-4o automated score for task decomposition) and Human Evaluation on a 5-point Likert scale, before and after calibration.
Calibration significantly reduces the difference between automated and human scores.}
    \label{tab:calibration_scores}
    \label{tab:calibration} 
\end{table*}

\begin{table*}[hbt!] 
    \centering
    \small

    \begin{tabular}{lp{0.8\linewidth}} 
    \toprule
       \textbf{Question} & Please list the phone numbers of the direct charter-funded schools that are opened after 2000/1/1.\\ \midrule
        \textbf{Evidence} & Charter schools refers to Charter School (Y/N) = 1 in the frpm\\ \midrule
\textbf{Gold SQL} & SELECT T2.Phone FROM frpm 
 AS T1 INNER JOIN schools AS T2 ON T1.CDSCode = T2.CDSCode WHERE T1.\textbf{Charter Funding Type} = "Directly funded" AND T1.Charter School (Y/N) = 1 AND T2.OpenDate > "2000-01-01" \\ \midrule 
\textbf{Predicted SQL} & SELECT T1.Phone FROM schools T1 JOIN frpm T2 ON T1.CDSCode = T2.CDSCode WHERE T1.\textbf{FundingType} = "Directly funded" AND T1.OpenDate > "2000-01-01" AND T2.Charter School (Y/N) = 1 \\ \bottomrule 
    \end{tabular}
    \caption{Example of an incorrect column name error in DeKeySQL's predicted SQL.
The model used "FundingType" from the 'schools" table alias T1, while the correct schema required "Charter Funding Type" from the "frpm" table alias T1 (or T2 if aliasing was different in gold).}
    \label{tab:column_error_example}
    \label{tab:column} 
\end{table*}

\begin{table*}[hbt!]
    \centering
    \small

    \begin{tabular}{lp{0.8\linewidth}}
    \toprule
       \textbf{Question} & Which school in Contra Costa has the highest number of test takers?\\ \midrule
        \textbf{Evidence} & ""\\ \midrule
\textbf{Gold SQL} & SELECT sname FROM satscores WHERE cname = "Contra Costa" AND sname IS NOT NULL ORDER BY NumTstTakr DESC LIMIT 1 \\ \midrule 
\textbf{Predicted SQL} & SELECT T1.sname FROM satscores T1 \textbf{JOIN schools 
 T2 ON T1.cds = T2.CDSCode} WHERE T2.County = "Contra Costa" AND T1.NumTstTakr IS NOT NULL ORDER BY T1.NumTstTakr DESC LIMIT 1 \\ \bottomrule 
    \end{tabular}
    \caption{Example of an incorrect aggregation error: DeKeySQL 
unnecessarily joined the 'schools" table when 'satscores" already contained sufficient information ("cname" for county, 'sname" for school name).}
    \label{tab:aggregation_error_example}
    \label{tab:aggregation} 
\end{table*}

\begin{table*}[hbt!]
    \centering
    \small

    \begin{tabular}{lp{0.8\linewidth}}
    \toprule
\textbf{Question} & How many test 
 takers are there at the school/s whose mailing city address is in Fresno?\\ \midrule
\textbf{Evidence} & ""\\ \midrule
\textbf{Gold SQL} & SELECT \textbf{T1.NumTstTakr} FROM satscores AS T1 INNER JOIN schools AS T2 ON T1.cds = T2.CDSCode WHERE T2.MailCity = "Fresno" \\ \midrule 
\textbf{Predicted SQL} & SELECT \textbf{SUM(T2.NumTstTakr)} AS totalTestTakers FROM schools T1 JOIN satscores T2 ON T1.CDSCode = T2.cds WHERE T1.MailCity = "Fresno" \\ \bottomrule 
    \end{tabular}
    \caption{Example of an incorrect operation error: DeKeySQL used 'SUM(T2.NumTstTakr)" when the question asked for the number of test takers at *each* school, implying individual counts, not a sum across all schools matching the criteria.}
    \label{tab:operation_error_example}
    \label{tab:Operation} 
\end{table*}

\begin{table*}[hbt!]
    \centering
    \small

    \begin{tabular}{lp{0.8\linewidth}}
    \toprule
\textbf{Question} & Name schools in Riverside which the average of average math score for SAT is greater than 400, what is the funding type of these schools?\\ \midrule
\textbf{Evidence} & \textbf{Average of average math = sum(average math scores) / count(schools).} \\ \midrule 
\textbf{Gold SQL} & SELECT T1.sname, T2.Charter Funding Type FROM satscores AS T1 INNER JOIN frpm AS T2 ON T1.cds = T2.CDSCode WHERE T2.District Name LIKE "Riverside\%" 
 GROUP BY T1.sname, T2.Charter Funding Type HAVING \textbf{CAST(SUM(T1.AvgScrMath) AS REAL) / COUNT(T1.cds) > 400} \\ \midrule 
\textbf{Gold SQL} & SELECT T1.sname, T2.FundingType FROM satscores AS T1 INNER JOIN schools AS T2 ON T1.CDSCode = T2.CDSCode WHERE T2.County = "Riverside" GROUP BY T1.sname, T2.FundingType HAVING \textbf{CAST(SUM(T1.AvgScrMath) AS REAL) / COUNT(T1.cds) > 400} \\ \midrule 
\textbf{Predicted SQL} & SELECT T1.sname, T2."FundingType" FROM satscores AS T1 JOIN schools AS T2 ON T1.cds = T2.CDSCode WHERE T2.County = "Riverside" GROUP BY T1.sname, T2."FundingType" HAVING \textbf{AVG(T2.AvgScrMath) > 400} \\ \bottomrule
    \end{tabular}
    \caption{Example of an incorrect evidence understanding 
error: DeKeySQL used "AVG(T2.AvgScrMath)" instead of following the provided evidence formula: "Average of average math = sum(average math scores) / count(schools)".}
    \label{tab:evidence_error_example}
    \label{tab:evidence} 
\end{table*}

\begin{table*}[hbt!]
    \centering
    \small

    \begin{tabular}{lp{0.8\linewidth}}
    \toprule
       \textbf{Question} & How many schools have an average SAT verbal score greater than 500 for students in grade 10? \\ \midrule
        \textbf{Evidence} & "" \\ \midrule
\textbf{Gold SQL} & SELECT COUNT(DISTINCT T1.sname) FROM satscores AS T1 JOIN schools AS T2 ON T1.cds = T2.CDSCode WHERE T1.AvgScrVerbal > 500 AND T2.GradeLevel = "High School" \\ \midrule
\textbf{Predicted SQL} & SELECT COUNT(T1.cds) FROM satscores AS T1 JOIN schools AS T2 ON T1.cds = T2.CDSCode WHERE T1.AvgScrVerbal > 500 AND T2.GradeLevel = "100" \\ \bottomrule
    \end{tabular}
    \caption{Example of an incorrect filtering error: DeKeySQL incorrectly used a numeric value "100" for "GradeLevel" which should have been a text value like "High School" or "Middle School" based on the database schema, indicating a mismatch between extracted keywords and actual database values.}
    \label{tab:filtering_error_example}
    \label{tab:Filtering} 
\end{table*}

\begin{figure*}[h]
    \centering
    \includegraphics[width=1\linewidth]{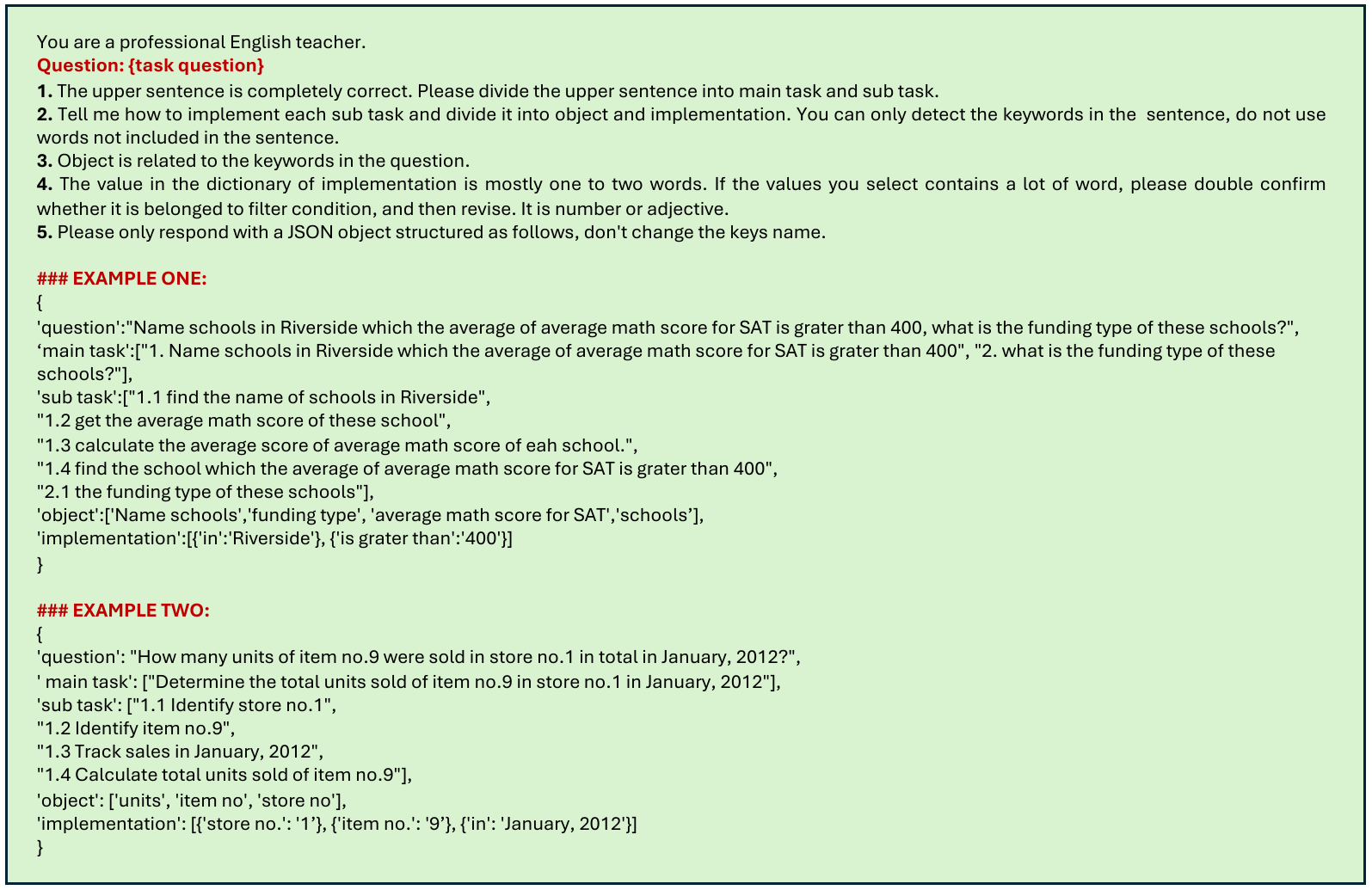}
    \caption{Prompt of keyword extraction and task decomposition.}
    \label{fig:nl2sql_keyword_extraction}
\end{figure*}

\begin{figure*}[h]
    \centering
    \includegraphics[width=1\linewidth]{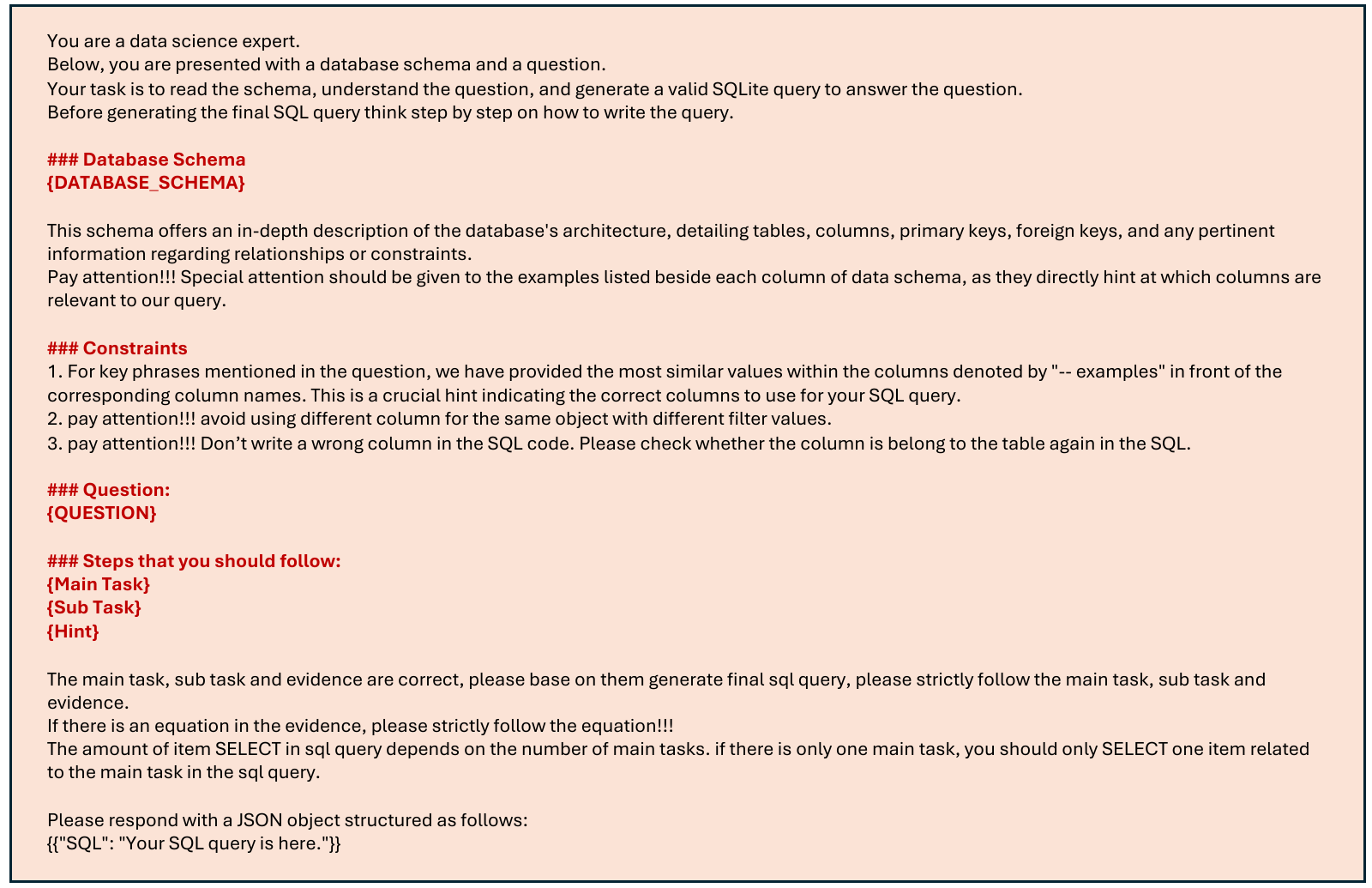}
    \caption{The prompt template used for the SQL Generation module within DeKeySQL. This template structures the input to the LLM, including database schema, user question, decomposed tasks, and extracted entities, guiding the model to produce an initial SQL statement.}
    \label{fig:candidate_generation}
\end{figure*}

\begin{figure*}[h]
    \centering
    \includegraphics[width=1\linewidth]{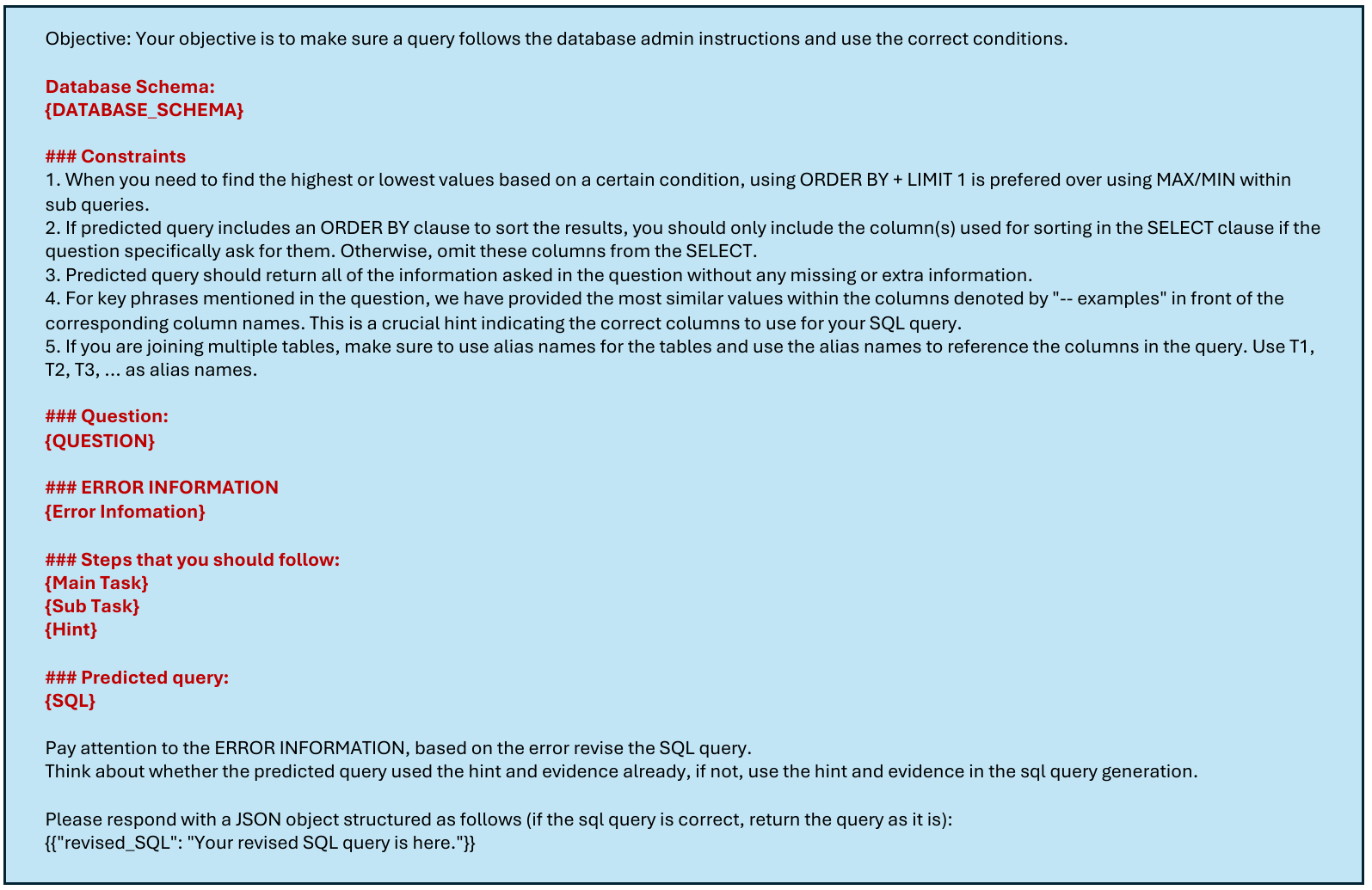}
    \caption{The prompt template used for the Revision module in DeKeySQL. This template provides the LLM with the erroneous SQL query and associated error messages, facilitating a targeted correction process to refine the SQL into a syntactically correct and operational query.}
    \label{fig:revision}
\end{figure*}

\begin{table*}[h]
\centering

\resizebox{0.8\textwidth}{!}{%
\begin{tabular}{p{0.2\textwidth} p{0.32\textwidth} p{0.3\textwidth} p{0.3\textwidth}}
\toprule
\textbf{Error Type} & \textbf{Question} & \textbf{Predicted} & \textbf{Ground Truth} \\
\midrule
\textbf{Miss Keyword} & Write the title and all the keywords of the episode that was aired on 3/22/2009. & object["title", "keywords"] & object["title", "keywords", "episode"] \\
\addlinespace 
\textbf{Wrong Keyword} & Write down the need statement of Family History Project. & object['statement"] & object["need statement"] \\
\addlinespace
\textbf{Useless Keyword} & List the tax code and inspection type of the business named "Rue Lepic". & object["tax code", "business", "inspection type", "name"] & object["tax code", "business", "inspection type"] \\
\bottomrule
\end{tabular}}
\caption{Qualitative Examples of Keyword Extraction Errors}
\label{tab:keyword_extraction_errors}
\end{table*}

\begin{table*}[h]
\centering

\resizebox{0.8\textwidth}{!}{%
\begin{tabular}{p{0.18\textwidth} p{0.2\textwidth} p{0.24\textwidth} p{0.24\textwidth}}
\toprule
\textbf{Error Type} & \textbf{Question} & \textbf{Predicted} & \textbf{Ground Truth} \\
\midrule
\textbf{Main Task Belongs to Sub Task} & What is the rental price per day of the most expensive children's film? & main task["1. Identify the most expensive children's film", "2. Find the rental price per day of that film"] & main task["1. Find the rental price per day of the most expensive children's film"] \\
\addlinespace
\textbf{Main Task Is Incomplete} & Which nation has the lowest proportion of people who speak an African language? Please state the nation's full name. & main task["Identify the nation with the lowest proportion of speakers of African languages"] & main task["1. Identify the nation with the lowest proportion of people who speak an African language", "2. State the full name of this nation"] \\
\addlinespace
\textbf{Sub Task Is Incomplete} & Please list the names of the male students that belong to the navy department. & main task["1. List the names of the male students that belong to the navy department"]\newline sub task["1.1 identify male students", "1.2 filter students belonging to the navy department"] & main task["1. List the names of the male students that belong to the navy department"]\newline sub task["1.1 find the male students", "1.2 filter students with navy department", "1.3 list the names of these male students"] \\
\addlinespace
\textbf{Sub Task Number Is Wrong} & For the business with great experience existed in Sun Lakes city, provide the user ID who gave review on it and user followers. & main task["1. Identify the business with great experience in Sun Lakes city", "2. Provide the user ID who gave review on it and user followers"]\newline sub task["1.1 identify the business with great experience in Sun Lakes city", "1.2 find the user ID of the person who gave review on this business", "1.3 get the number of user followers for this user"] & main task["1. Identify the business with great experience in Sun Lakes city", "2. Provide the user ID who gave review on it and user followers"]\newline sub task["1.1 find the business with great experience in Sun Lakes city", "2.1 identify the user ID who gave review on this business", "2.2 find the followers of these users"] \\
\addlinespace
\textbf{Sub Task Is Ambiguous} & Is SuperSport Park located at Centurion? & main task["1. Is SuperSport Park located at Centurion?"]\newline sub task["1.1 verify the location of SuperSport Park"] & main task["1. Is SuperSport Park located at Centurion?"]\newline sub task["1.1 find the location of SuperSport Park", "1.2 check if the location is at Centurion"] \\
\bottomrule
\end{tabular}}
\caption{Qualitative Examples of Task Decomposition Errors}
\label{tab:task_decomposition_errors}
\end{table*}

\end{document}